\pdfoutput=1
\documentclass[letterpaper, 10 pt, journal, twoside]{IEEEtran}
\usepackage{color}
\usepackage{float}
\usepackage[subrefformat=parens,labelformat=parens]{subfig}
\usepackage[ruled,vlined]{algorithm2e}
\usepackage{threeparttable}
\usepackage{url}

\usepackage[pdftex]{graphicx}
\usepackage{amsfonts}
\usepackage{graphics} %
\usepackage{amsmath} %
\usepackage{amssymb}  %
\usepackage{enumitem}  %
\usepackage{array}
\usepackage{caption}
\usepackage{cite}
\usepackage{xcolor}

\usepackage[colorlinks,linkcolor=black,anchorcolor=black,citecolor=black,urlcolor=black]{hyperref}
\usepackage{balance}

\graphicspath{{./fig/},{./logo/},{../logo/},{./pic/}}
\begin{document}

\title{GPU-accelerated Incremental Euclidean Distance Transform for Online Motion Planning of Mobile Robots}

\author{Yizhou Chen,~Shupeng Lai,~Jinqiang Cui,~Biao Wang,~Ben~M.~Chen%

\thanks{Manuscript received: January 23, 2022; Revised: April 11, 2022; Accepted: May 8, 2022. This paper was recommended for publication by Editor Javier Civera upon evaluation of the Associate Editor and Reviewers’ comments. This work was supported in part by the Research Grants Council of Hong Kong SAR (Grant No: 14209020) and in part by the Hong Kong Centre for Logistics Robotics. \textit{(Corresponding author: Shupeng Lai.)}

Yizhou Chen and Ben M. Chen
	are with the Department of Mechanical and Automation Engineering,
	The Chinese University of Hong Kong,
	Shatin, N.T., Hong Kong 999077
	{ (e-mail: yzchen@mae.cuhk.edu.hk; 
	bmchen@cuhk.edu.hk)}.
	
Shupeng Lai is with the Department of Electrical and Computer Engineering, National University of Singapore, Singapore 117583 	{(e-mail: shupenglai@u.nus.edu)}.
	
Jinqiang Cui is with Peng Cheng Laboratory, Nanshan, Shenzhen, Guangdong, China 518055 {(e-mail: cuijq@pcl.ac.cn)}.

Biao Wang is with College of Automation Engineering, Nanjing University of Aeronautics and Astronautics,
	Nanjing, Jiangsu, and Peng Cheng Laboratory,
	Nanshan, Shenzhen, Guangdong, China 518055
	{(e-mail: wangbiao@nuaa.edu.cn)}.
}
}

\markboth{IEEE Robotics and Automation Letters. Accepted May, 2022}
{Chen \MakeLowercase{\textit{et al.}}: GPU-accelerated incremental Euclidean distance transform for online motion planning of mobile robots}

\maketitle

	\begin{abstract}
In this letter, we present a volumetric mapping system that effectively calculates Occupancy Grid Maps (OGMs) and Euclidean Distance Transforms (EDTs) with parallel computing. 
Unlike these mappers for high-precision structural reconstruction, our system incrementally constructs global EDT and outputs high-frequency local distance information for online robot motion planning. 
The proposed system  constructs OGM with massive amount of data from multiple types of sensor inputs. Using GPU programming techniques, the system quickly computes EDT in parallel within local volume. The new observation is continuously integrated into the global EDT using the parallel wavefront algorithm while preserving the historical observations.  
Experiments with datasets have shown that our proposed approach outperforms existing state-of-the-art robot mapping systems  and is particularly suitable for mapping unexplored areas. In its actual implementations on aerial and ground vehicles, the proposed system achieves real-time performance with limited onboard computational resources.	
	\end{abstract}
	
\IEEEpeerreviewmaketitle

\section{Introduction}
\IEEEPARstart{A}{n} accurate representation of the local environment is critical to the effectiveness of mobile robot navigation \cite{lan2021survey,mondal2021offline,yuan2021survey,cui2016search}. Especially, the distance information to nearby obstacles is a must in many motion planning algorithms \cite{ratliff2009chomp,zhou2019robust, arora2021environment}. 
To that end, OGMs and EDTs are often used to bridge  environment representation and planning. Both the OGM and EDT are voxel maps. In an OGM, each voxel records the probability of being occupied by an obstacle. An EDT records the distance information from each voxel to the nearest obstacle. Typically, a local EDT is required to be updated at high frequency for real-time performance of the local planner, while a global EDT is queried less frequently by the global planner.

Creating these maps is challenging due to limited onboard computational resources and real-time requirements of motion planning. To solve this problem, many researchers (see, e.g., \cite{ jarvis1993distance,lau2010improved,oleynikova2017voxblox, han2019fiesta}) choose to sacrifice precision for increased runtime speed. However, their wavefront-based methods can incur large errors in real-world scenarios because the distance values only propagate on observed voxels. Besides, computational resources are highly stretched when updating global EDT with a lightweight onboard computing unit.
As the demand for a faster update procedure of global incremental EDT rises, it is natural to explore the possibility of computing OGM and EDT in a parallel fashion using GPU onboard. Unfortunately, most of the methods designed for CPUs reported in the literature have very limited parallelization capability.

    We also note that various approaches to distance transformation have been studied in the pattern recognition society (see, e.g., \cite{saito1994new,meijster2002general,Cao2010Parallel}). These methods are mainly developed for fast distance transformation on a fully known binary map, usually a 2D image. However, in robotic applications, the environment map needs to be incremental as the robot moves, while the standard sensors of mobile robots have a limited sensing range. Moreover, the size of the map cannot be determined in advance for situations in unknown environments.

    Motivated by the fast demand in real-time applications for small robots, such as small UAVs and UGVs, we propose in this letter an innovative and effective approach to increase the efficiency and accuracy of the mapping process through massive-scale parallelization. It is designed for the navigation purposes of a mobile robot and allows an incremental map to be constructed with dynamically adjustable map size.
The proposed mapping system is versatile enough to construct OGMs from commonly used onboard sensors such as depth cameras and 2D- or 3D-LiDAR. It computes a fast and exact batch EDT with the parallel Euclidean distance transform. The batch EDT is then integrated into the global EDT through a parallel wavefront mechanism. With the utilization of parallel computing techniques, both the local and global EDT can be updated in real-time. The proposed system can be deployed on miniature computing devices with GPU, such as Nvidia  Xavier NX, commonly used for small robots or unmanned systems. We will make the source code of our system available  as an open-source project.\footnote{https://github.com/JINXER000/GIE-mapping}

The contributions of this work are: i)
the development of an efficient approach that generates OGM and EDT through massive parallelization;
ii) the design of an EDT integration method that allows the incremental construction of global EDT with dynamically adjustable map size; and lastly iii)
the integration of the proposed algorithm in various ground/aerial mobile robots to demonstrate its effectiveness with onboard motion planning. 

The rest of this letter is organized as follows: 
In Section~\ref{relate}, we recall some background materials in fast EDT algorithms and state-of-the-art mappers. In Section~\ref{framework}, we present the overview and framework of our proposed mapping  system. The technical details of the proposed system are given in Section~\ref{chap alg}. The experimental and simulation results that demonstrate the effectiveness of our system are presented in Section~\ref{chap res}, while the result conducted on robotic platforms is given in
Section~\ref{newsection}. Finally, we draw some concluding remarks in Section~\ref{Last_Sec}. 

\section{Related Work and Background}\label{relate}

 In occupancy maps, such as OGM or Octomap, they are usually discretized into voxel grids that store occupancy information.  Recently, some researchers (see, e.g.,  \cite{vespa2018efficient,funk2021multi}) use occupancy maps to perform surface reconstruction and motion planning at the same time. They have mainly focused on high-resolution space representation rather than fast onboard distance checking. Although the constructed maps can be used in collision checking in corridor-based planners, information on the distance to obstacles is generally not captured.  

{For many optimization-based motion planning algorithms, it is necessary to query the distance between the trajectory and its nearest obstacle. The demand of multiple queries in each planning iteration  dramatically affects the overall efficiency of the planning algorithm.  One common approach is to pre-compute an EDT map  such that the query process can be executed in $O(1)$ time.	
For robots, such as miniature aerial vehicles (MAVs), the onboard computers generally have limited computational power due to the very limited payload of the vehicle. The demand on real-time EDT generation has forced to exclude EDT mappers in the process (see, e.g., \cite{jarvis1993distance,wu2021faithful}). Similarly, researchers in \cite{zhou2019robust,ratliff2009chomp}  only maintain a local and memoryless EDT to represent the environment around the vehicle, whereas obstacles outside the local area are ignored.
Also, the works in \cite{oleynikova2017voxblox, han2019fiesta} construct incremental EDT online by careful implementation of dedicated data structures. 
More specifically, 
the method, Voxblox in \cite{oleynikova2017voxblox} builds EDTs out of the so-called truncated signed distance fields (TSDFs), which contains projected distance information within a truncation radius near the surface. They utilized the distance to propagate throughout the whole space in a quasi-Euclidean metric and finally obtain an EDT map. However, the resulting EDT map has two sources of errors, which would accumulate through the propagation. 
On the other hand, the approach, FIESTA in \cite{han2019fiesta}, uses double-linked lists (DLLs) to record distance information and improve both mapping precision and efficiency.  Although their methods have been implemented on mobile robots with vision-based sensors, the computation cost is extremely high when maintaining a global EDT in large areas.}

Many studies have been carried out in computational geometry and pattern recognition society in developing fast and exact distance transform algorithms. Saito and Toriwaki \cite{saito1994new} devise an algorithm to produce EDT by independently scanning on each dimension. 
Meijster \textit{et al.}  \cite{meijster2002general} follow the concept of \cite{saito1994new} to enhance the overall performance by additionally utilizing properties of parabola intersections.
These decomposition-based distance transform algorithms break down the whole mapping procedure into strips or structuring functions, which is suitable for parallel computing. 
 However, these methods compute EDT in a  batch of fixed-size memory, which is often not the case in robotics applications.  To address these issues, we employ the technique of voxel hashing \cite{niessner2013real} to store the mapping history and further utilize  parallel wavefront to maintain the consistency between the new incoming data and the existing map. 
 {Finally, we would like to note that our previous work, i.e., the technique, GPU-LO in \cite{chen2021gpu}, only considers the limited observation problem during the mapping process. We propose in this letter more wavefronts in the system proposed to tackle problems associated with real-world mapping tasks.}

\section{Overview of the Proposed System Framework} \label{framework}

\begin{figure}[htb]
	\centering
	\includegraphics[width=0.5\textwidth]{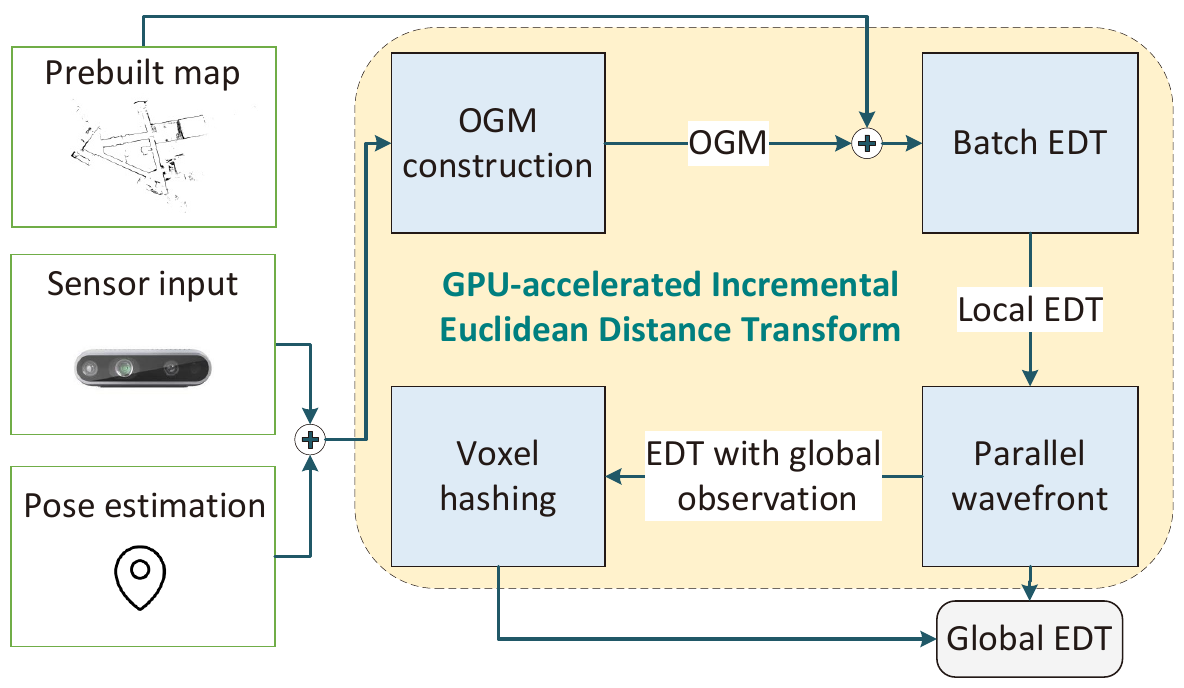}
	\caption{Framework of the proposed system}
	\label{fig:sys}
\end{figure}

In what follows, we present a framework,  depicted in Fig.~\ref{fig:sys}, for generating OGM and EDT through massive parallelization, which can be divided into three stages: OGM construction, batch EDT, and global EDT integration. 

In the OGM construction stage, the system intakes data from the sensors and builds the OGM. Position and attitude information are usually estimated by localization algorithms, while range measurement can be obtained from onboard range sensors. {The range measurement is then integrated into the global OGM as occupancy probabilities by parallel implementations of ray casting and volumetric projection.} 

In the batch EDT stage, the parallel EDT algorithm transforms the OGM in the Euclidean distance metric. Domain knowledge, such as pre-built OGM or pre-defined forbidden zones,  can also be incorporated into the EDT process.  The algorithm scans each dimension in turn and calculates the distance value for each voxel.

Finally, in the global EDT integration stage, the EDT in the local area is integrated into the hash table that stores global EDT. After identifying the proper sources, the parallel wavefront is utilized to propagate  actual distance values over both the local and global EDT. Hence the batch EDT is updated with global observation, and the history can be preserved  in the hash table. 

\section{Algorithms and the Pipeline} 
\label{chap alg}
We present in the following the detailed algorithms for constructing OGM, batch EDT and global EDT integration.
\subsection{{OGM Construction}} 
\label{sect map const}

Two GPU-accelerated algorithms for OGM construction are developed. The first one is a massively-parallel version of the traditional ray casting algorithm \cite{amanatides1987fast}, while the second one is a parallel volumetric projection method utilizing sensor models \cite{curless1996volumetric}.
 Each voxel of OGM contains its occupancy probability and its visibility information. 

In ray casting, given a pixel at entry $(i,j)$ of a depth image, the algorithm first transforms the pixel to a 3D voxel $\mathcal{V}_{x, y, z}$:
\begin{equation}
\mathcal{V}_{x, y, z}=\mathcal{T}^{-1}K^{-1}\mathcal{D}_{i,j,w}, \label{eq:back-proj}
\end{equation}
where $\mathcal{T}$ is the transformation from the world frame to the sensor frame, $K$ is the intrinsic parameters of the sensor, and $\mathcal{D}_{i,j,w}$ is the measurement at $(i,j)$ with depth $w$.  
 Rays are then cast from the sensor to measured obstacle points and update the occupancy probability of each voxel along the line. Denoting the number of points perceived as $n$, the average travel distance as $l$, and the voxel size as $q$, the computational complexity of  ray casting process is in the order $O(nl/q)$. 
 One most significant challenge to parallelize ray casting is that multiple rays might try to write to the same voxel with different data, leading to serious synchronization issues of possible memory conflict. We resolve this problem using CUDA-atomic operations, which forces I/O operations to be mutually exclusive. 

On the other hand, we adopt the volumetric projection method of \cite{curless1996volumetric}, which is developed to construct TSDF using optical triangulation scanners, to build OGMs with sensors such as 2D- or 3D-LiDAR. Instead of casting rays from pixels in the depth image to the 3D space, the projection  computes the appropriate sensor value of each voxel in the local volume by simulating the acquisition process according to the sensor model. Assuming the range measurement data to be two-dimensional, the projection from the voxel $(x,y,z)$ to the sensor data entry $(i,j)$ can be formulated as:
\begin{equation}
\mathcal{D}_{i, j, \bar{w}}=K\mathcal{T}\mathcal{V}_{x, y, z},
\end{equation}
where $\mathcal{D}_{i, j, \bar{w}}$ is the projection onto the sensor with the projected range $\bar{w}$.
Thereafter, by comparing the projected range $\bar{w}$ with the measured range $w$, occupancy information can be updated. The detailed procedure can be found in our previous work \cite{chen2021gpu}. In volumetric projection, the computational complexity is proportional to the volume size, i.e., $O(N)$, if there are $N$ voxels within the local volume. As $N$ is generally huge in many applications, such an approach is rarely used in CPU. It can be massively parallelized, however, in GPU without CUDA-atomic operations. 

{The choice among various OGM construction methods depends heavily on input data. For example, during the construction of point cloud, some sensors choose not to generate a point if the corresponding ray hits no object within the detection range. If a ray casting method is applied directly,  the volume covered by such a ray will not be updated as known and obstacle-free. Besides, when the resolution of the range sensor is extremely high, a down-sampling procedure is required to speed up ray casting. As such, we give preference to volumetric projection if intrinsic parameters of the range sensor can be obtained.} 

\subsection{Batch EDT} \label{sec map percept}
{Unlike the existing EDT construction methods of \cite{lau2010improved, oleynikova2017voxblox,han2019fiesta}, which propagate distance from newly observed voxels in an ordered fashion, our method first evaluates EDT in the local volume with a GPU-based scanning technique without introducing propagation error.
 The parallel batch EDT  workflow is given in Algorithm~\ref{alg locedt}, which consists of one initial scan and several auxiliary scans. 
We extend the scanning method of Meijster \textit{et al.} \cite{meijster2002general},
which performs exact EDT in linear time, to the 3D space.
It decomposes memory into independent lines and can therefore be parallelized.
In fact, any other EDT algorithm with similar attributes {(see, e.g., \cite{Cao2010Parallel})} 
 can be applied in our framework. In addition to calculating distance values, our algorithm tracks the nearest obstacle of each voxel during scans, which is essential for subsequent wavefront propagation.
 }

\begin{algorithm}\footnotesize
\caption{BATCHEDT}\label{alg locedt}
\LinesNumbered %
\KwIn{OGM $O$, dimension of OGM $dim$}%
\KwOut{Batch EDT  $E$ in the local volume and  nearest obstacles recording $P$ }%
 $ d \gets 1$\;
 $ E, P \gets \mathsf{INITSCAN}(O)$\;
 $ E \gets \mathsf{TRANSPOSE}(E,d)$\;
\While{$d \leq dim$}{ 
  $ d \gets d+1$\;
$ E, P \gets \mathsf{AUXSCAN}(E, P)$\;
 $ E \gets \mathsf{TRANSPOSE}(E,d)$\;

}

$ E \gets \mathsf{CHECKVALID}(E, O)$\;
\Return $E, P$
\end{algorithm}

{In the case of computing a 3D EDT, our objective is to find the nearest obstacle $(i, j, k)$  with respect to each voxel $(x,y,z)$.}
 In the initial scanning INITSCAN(), distance values of all occupied voxels are set as 0, while those of empty voxels are set as \textit{inf}. 
When generating EDT in 1D, each column is required to be swept forward and backward to compute each grid of its distance to the nearest obstacle within the same column. The sweep always starts with a value of 0 at the obstacle and then increments in empty areas. This procedure can be described as:
\begin{equation} \label{2sweep}
{ G(x, y, z)=\min_{j \in rows}\{\left|j-y\right||\ G(x,j,z)=0\} ,}
\end{equation}
where $G(x, y, z)$ is the  distance value of voxel $\mathcal{V}_{x, y, z}$ when scanning the first dimension.

\begin{figure}[htb]
	\centering
	\includegraphics[width=0.56\textwidth]{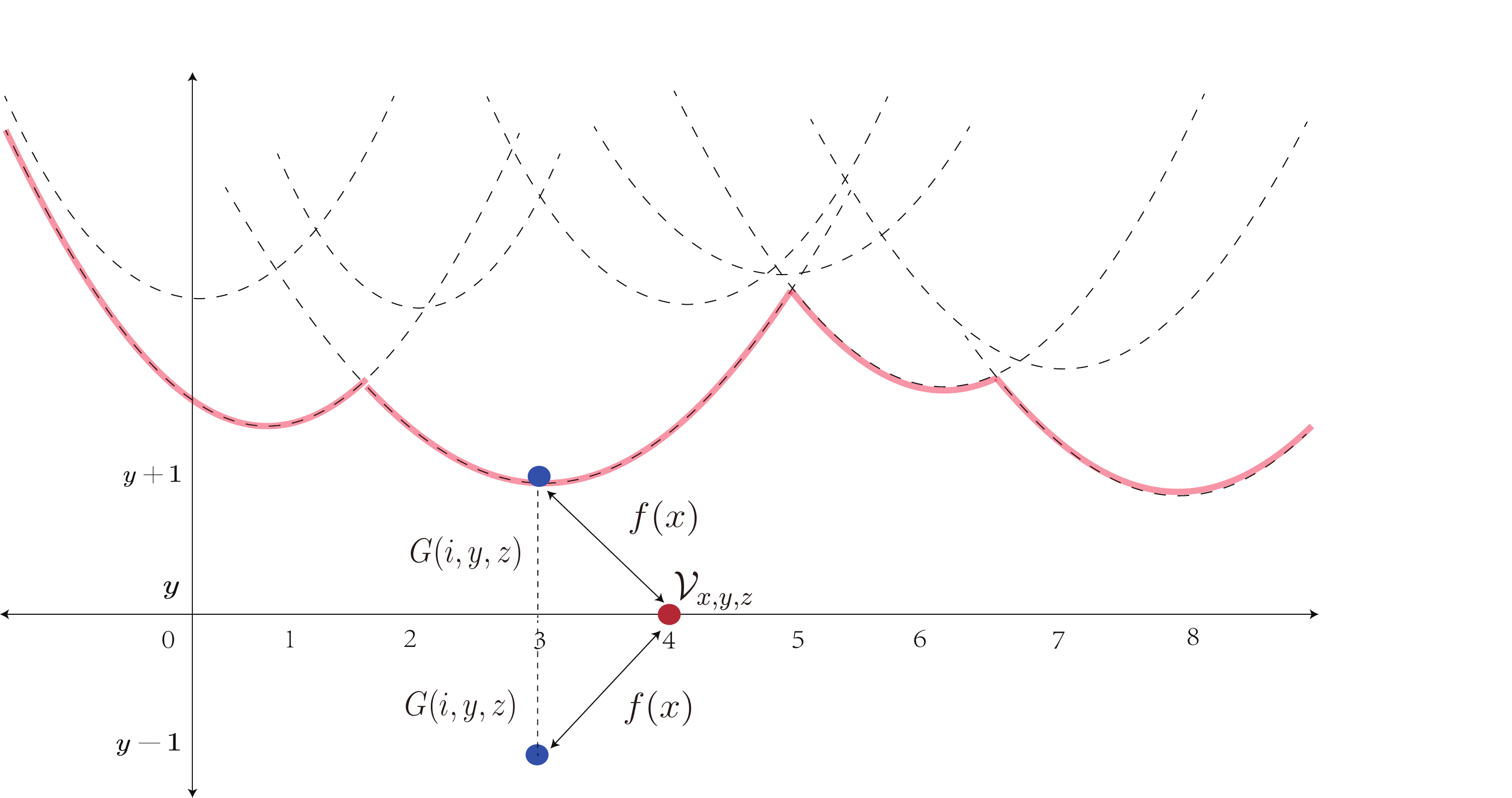}
	\caption{An illustration of distance functions in AUXSCAN(). When a {voxel $\mathcal{V}_{x, y, z}$}  (the red dot) is inspected by the current thread, it will be assigned to one parabola segment given the lower envelope (pink curve). The blue dots denote the possible positions of the nearest obstacle of $\mathcal{V}_{x, y, z}$.}
	\label{fig:parabola}
\end{figure}

Fig.~\ref{fig:parabola} illustrates the auxiliary scan  on the X-Y-plane. Every thread
In the auxiliary scan  on the X-Y-plane, every thread
executes a sweeping process along X-axis while fixing the value on Y- and Z-axes. Assuming the nearest obstacle of  voxel $\mathcal{V}_{x, y, z}$ lies on the column $C_i$, the distance  function of its $x$ value can be written as
\begin{equation}\label{vox dist}
 {f(x)=(x-i)^2+G(i,y,z)^2       ,} 
\end{equation}
where  $G(i,y, z)$ is regarded as a constant since it is already calculated at INITSCAN(). The distance function of each voxel can be represented as a parabola shifted by its $G$ value (as shown in Fig.~\ref{fig:parabola}). Essentially, the calculation of  distance values is  to find the lower envelope of the union of all parabolas, which can be formulated as
\begin{equation}\label{map dist}
{ F(x,y, z)=\min_{i \in cols}\{(x-i)^2+G(i,y, z)^2\} ,  }   
\end{equation}
where $cols$ denotes all columns of the map in the X-Y-plane.
When traversing the row,  parabolas are inserted one by one. If the incoming parabola intersects the current lower envelope, a new parabola segment will be created, and the lower envelope will be updated. 
We refer readers to   \cite{meijster2002general} for details.
After calculating distance values on the X-Y plane, the function AUXSCAN()  can then be adopted to progress on Z-axis in a similar fashion. Finally, the nearest obstacles in the entire 3D space are captured.

Lastly, each scan is to be followed by a TRANSPOSE() function. We utilize a tensor transpose library, cuTT\cite{hynninen2017cutt}, to reorder the memory of the local volume so that coalescing memory access  can be achieved. After all  scans, each voxel of the EDT goes through validation CHECKVALID(). {Note that during the OGM construction, voxels are divided into three categories: free, occupied, and unknown. A voxel will be further integrated into the global EDT only if it is ever observed by the robot.} 

\subsection{Global EDT Integration} \label{chap global EDT integration}

\begin{figure}[htbp]  %
    \centering    %
    
    \subfloat[] %
    {
        \begin{minipage}[t]{0.23\textwidth}
            \centering          %
            \includegraphics[width=0.9\textwidth]{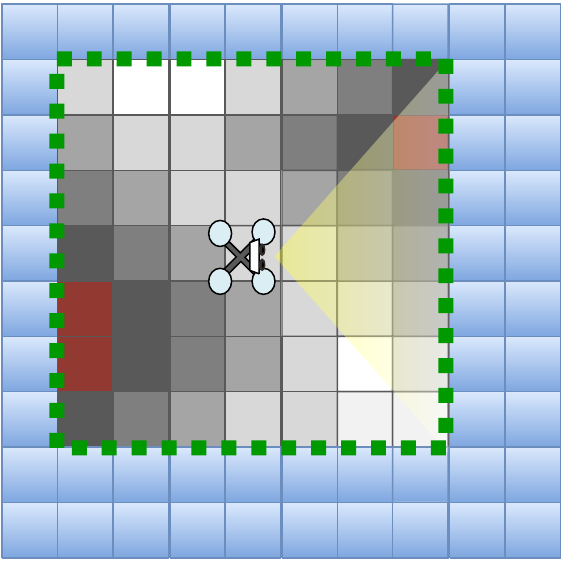}   %
        \end{minipage}%
        \label{limit-observe-a}
    }%
    \subfloat[] %
    {
        \begin{minipage}[t]{0.23\textwidth}
            \centering      %
            \includegraphics[width=0.9\textwidth]{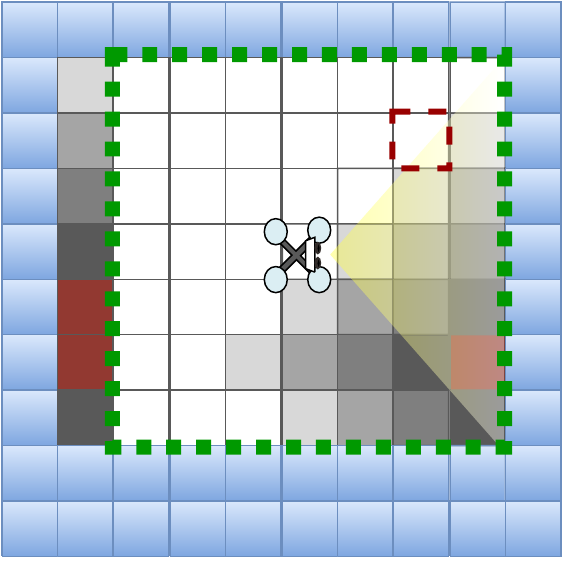}   %
        \end{minipage}
        \label{limit-observe-b}
    }%
    
    \caption{Limited observation and dynamic obstacle problems. Batch EDT is computed within green dotted line. The obstacles are painted in the dark red grids. The color shade in the observed space represents the distance value. The blue grids are unobserved area. 
    } %
    \label{limit-observe}  %
\end{figure}

With the batch EDT process, the observable region in the local volume is filled with EDT at each timestamp. 
{However, the mapping accuracy is greatly affected by dynamic obstacles, sensor noise, and more severe, limited observation, which can be illustrated in Fig.~\ref{limit-observe}. More specifically, Fig.~\subref*{limit-observe-a} depicts an MAV flying in 2D grids. When computing batch EDT in the local volume (green dotted-line box), the map only reveals the information inside the local volume and has no knowledge about sensed regions in history. In Fig.~\subref*{limit-observe-b}, as the vehicle moves to the right, the obstacles at the left boundary are no longer inside the updating range and are ignored in the batch EDT process. In consequence, inconsistencies between the global EDT and the current local EDT occur. Besides, dynamic obstacles can also pop up (the light salmon grid at lower right) and disappear (the grid with red dashed line at upper right) in the field of view.}  

{To tackle the above problems, we combine the batch EDT with our global integration algorithm. Essentially, this algorithm extract frontiers (Algorithm~\ref{alg getsource}), and then perform parallel wavefront (Algorithm~\ref{alg parbfs}) separately. Afterwards, the updated EDT in the current local volume is stored in a GPU hash table \cite{niessner2013real}, which aims to make the mapping spatial efficient. The updated portion of hash table on GPU is streamed to CPU; hence the global EDT on both CPU and GPU are continuously constructed in the run time.}

{The source extraction function GETSOURCE() of Algorithm~\ref{alg getsource} is explained in detail. Line~\ref{algl:bdr} inspects each voxel {\it cur} that lies on the boundary of the batch EDT and its neighbor {\it nbr} that lies outside the local volume. If the nearest obstacle {\it nbr.parent} is inside the local volume, but the distance value of the neighbor in GPU hash table DIST($GHash,nbr$) can be raised by one obstacle inside the local volume, it implies that {\it nbr.parent} disappears in the robot’s field of view(Line~\ref{algl:disappear}). Otherwise, if an obstacle suddenly pops up in the new observation , it will lower the distance value of {\it nbr} (Line~\ref{algl:popup}). If the nearest obstacle {\it nbr.parent}, which is outside the local volume, can lower the distance value inside the local volume $\mathsf{DIST}(E,cur)$, then {\it cur} is appended to the frontier corresponding to the limited observation situation (Line~\ref{algl:limit}).}

\begin{algorithm}\footnotesize
\caption{GETSOURCE}\label{alg getsource}
\LinesNumbered %
\KwIn{Current local EDT $E$, global EDT stored in GPU hash table $GHash$}%
\KwOut{Initial frontiers of propagation}%
 $frontierA,frontierB,frontierC \gets \varnothing$ \\
 \ForAll{$cur \in E.boundary$}{ \label{algl:bdr}
   \ForAll{$nbr \in \{x | x \in cur.nbrs \wedge	x \notin E.volume \}$}{

\uIf(){$||cur.parent - nbr||_{2} > \mathsf{DIST}(GHash,nbr)$ $\wedge$ $nbr.parent \in E.volume$}{
$nbr.raise \gets \mathrm{true}$ \\
$frontierA \gets frontierA \cup \{nbr\} $  \label{algl:disappear} \\
    
}
\uElseIf(){$||cur.parent - nbr||_{2}  < \mathsf{DIST}(GHash,nbr)$ }{
$frontierB \gets frontierB \cup \{nbr\}$    \label{algl:popup}
}
    \uIf(){$||nbr.parent - cur||_{2} < \mathsf{DIST}(E,cur) $ $\wedge$ $nbr.parent \notin  E.volume$}{
         $frontierC \gets frontierC \cup \{cur\}$  \label{algl:limit}
}

   }
   }
   \Return $frontierA, frontierB, frontierC$
\end{algorithm}

\begin{algorithm}\footnotesize
\caption{PARWAVEFRONT (raise)}\label{alg parbfs}
\LinesNumbered %
\KwIn{Initial frontiers  $frontierA,$ $ frontiersB$, current local EDT $E$, and global EDT in GPU hash table $GHash$}%
\KwOut{Updated local EDT and global EDT}%

\While{$frontierA.size \neq 0$}{ \label{outgrow}
$ \mathsf{CHOOSELEVEL}(frontierA)$\\ \label{change level}
    \ForAll{$cur \in frontierA$}
    {
      $frontierA \gets frontierA \backslash \{cur\} $ \\ \label{algl:cur}
    \ForAll{$nbr \in \{x|x \in cur.nbrs \wedge x.raise = \mathrm{false}  \}$}{ \label{algl:nbr}
     \If{$nbr.parent \in  E.volume \wedge \mathsf{DIST}(E,nbr.parent) \neq 0$}{
     $\mathsf{DIST}(GHash,nbr) \gets ||cur.parent - nbr||_{2}$ \\
     $nbr.parent \gets cur.parent $ \\
     $nbr.raise \gets \mathrm{true}$ \\
     
     $frontierA \gets frontierA \cup \{nbr\}$
     }
      \ElseIf{$\mathsf{DIST}(GHash,cur) > ||nbr.parent - cur||_{2}$}{ \label{algl:fb}
            $frontierB \gets frontierB \cup \{cur\}$
        }
    }
    }
  
}
 \Return $E$, $GHash$
\end{algorithm}

{ 
Subsequently, the parallel wavefront function PARWAVEFRONT() in Algorithm~\ref{alg parbfs} propagates the accurate distance from the frontiers to both inside and outside the local volume. The essential problem of conducting propagation in parallel is that GPU implementations of wavefront algorithms \cite{harish2007accelerating, deng2009taming} are inherently computationally inefficient and can even be slower than traditional CPU-based algorithms. We tackle this problem by extending hierarchical frontiers \cite{luo2010effective}, which originally solves single-source shortest path problems, to our wavefront context, where a voxel can be visited more than once in one or more iterations. Algorithm~\ref{alg parbfs} shows the details of the parallel wavefront propagation to raise the distance value caused by vanished obstacles. In the function CHOOSELEVEL(), the frontier levels are changed in accordance with the number of elements of the frontier. If it exceeds the maximum size of a GPU thread-block, the frontier is stored on the global memory for multiple thread-blocks to access. Otherwise, they are placed in the shared memory of one thread-block, which is faster than on the global memory. 
In Line~\ref{algl:cur}, each cell {\it cur} of the frontier is placed on a single thread. Starting at Line~\ref{algl:nbr}, we enumerate all the neighbors of {\it cur} whose distance value is not raised. If the nearest obstacle of {\it nbr} is found disappeared inside the local volume, {\it nbr} will be propagated by {\it cur} and be inserted back into {\it frontierA}. Otherwise, in Line~\ref{algl:fb}, if the current voxel encounters a neighbor that has a closer obstacle to {\it cur}, the algorithm adds {\it cur} to another frontier {\it frontierB} for a lowering wavefront. The algorithm for lowering wavefront is similar to Algorithm~\ref{alg parbfs} except that it propagates the minimum distance value to the neighborhood. Following the wavefront processing {\it frontierB} outside the local volume, the wavefront of {\it frontierC} begins to propagate global observation into the local volume. We omit this algorithm in the paper because of page limitation. Interested readers are referred to the detailed algorithm appended in the supplementary material\footnote{https://github.com/JINXER000/GIE-mapping/blob/main/doc/sup-mat.pdf}. Finally, the propagation ends once all frontiers become empty.}

\section{Experimental Results} \label{chap res}
In this section, we conduct a series of thorough tests on three datasets: the UAV-pillar dataset, the UGV-Corridor dataset, and the Cow-and-Lady dataset.  
The UAV-Pillar dataset is synthesized in the Gazebo simulator, where an MAV is flying through pillars at a constant speed, gathering range information with an onboard 3D LiDAR. The UGV-Corridor dataset is recorded in a real-world environment with long and narrow corridors, where a UGV is equipped with a 16-line Velodyne LiDAR and drives through pedestrians. 
The Cow-and-Lady dataset is presented with the work Voxblox \cite{oleynikova2017voxblox}, and its data is collected using a depth camera in a small room. 

\subsection{Simulation in An Unexplored Environment} \label{sec changerate}
While exploring an unknown space, the robot updates its map with continuous sensor data stream. The degree of change in the sensed environment also affects the computational efficiency of the mapping process. We demonstrate that the speed of our method is stable while exploring an unknown environment. 

\begin{figure}[htb]
	\centering
	\includegraphics[width=0.38\textwidth]{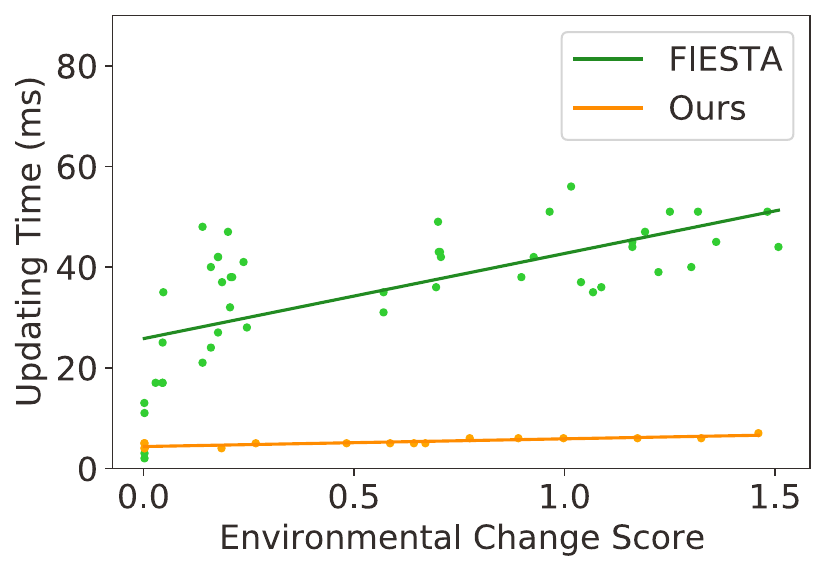}
	\caption{EDT operation speed  with different environmental change score.}
		\label{fig:change_rate}
\end{figure}

To quantify the difference between two sensed environments, we define the change score $V$ as:
\begin{equation}
    V(t_{1}, t_{2}) =  \mathop{\mathbb{E}}_{i\in S}(|E_ {t_1}(i)-E_{t_2}(i)|), 
\end{equation}
where $S$ is the set of all grids in the map,  $i$ is the index of the grid,  and $E_{t_1} , E_{t_2}$ are respectively the ground truth EDTs  at two distinct timestamps $t_{1}, t_{2}$ over the global range. The unknown volume is treated as empty during ground truth EDT calculation. If there is no difference between the sensed environment at $t_{1}$ and $ t_{2}$, the score will be 0. The more the sensed environment changes, the higher the score is. The EDT updating speed versus the environment change score can be found in Fig.~\ref{fig:change_rate}, where the resolution is set as 0.2~m and the local update range is 20~m~$\times$~20~m~$\times$~6~m.  The figure shows that the EDT construction time continues to increase as the environmental change score increases. However, the construction speed of our mapper remains almost constant. This is because the traditional mappers construct EDT purely with wave propagation and rely more on existing EDT.

\subsection{Experiments on Real-world Datasets} \label{sec dataset}

We test each stage of our system separately to show how differences in techniques that affect performance. First, we show the necessity of transpose operation in the batch EDT stage. The coalescing pattern is realized by forcing each thread to access adjacent memory, and  scanning is significantly accelerated. In this experiment, we test the UGV-Corridor dataset on an Xavier NX. Table~\ref{tab:coalesced} shows the difference between the approaches with and without the coalesced memory access. It is observed that the parallel EDT operation is much faster in the coalescing case.
\begin{table}[htbp]
		\centering
		\footnotesize
		\caption{Batch EDT time with and without coalesced memory access}
		\label{tab:coalesced}
		\tabcolsep 3pt %
		\renewcommand\arraystretch{1.2}
		\begin{threeparttable}
\begin{tabular}{c c}
\hline \hline
\textbf{Method}          &  \textbf{Average Updating Time (ms)} \\\hline
Original & 46 \\
Coalescing & \textbf{19} \\\hline \hline

\end{tabular}
\end{threeparttable}
\end{table}

Apart from the global EDT integration technique proposed in Section~\ref{chap global EDT integration}, we evaluate the methods on the Cow-and-Lady dataset with different graph traversal algorithms. The experiments are 
done on a PC with Intel i5-8300H CPU and a GTX1060 GPU, mapping 6~m~$\times$~6~m~$\times$~3~m local range with 0.05~m resolution. {In our method, the GPU thread-block limit is set as 512, and 6-connectivity is used.}  The results of average updating time are shown in Table~\ref{tab:diff-prop}. The baseline solution using dynamic brushfire on CPU with the bucket-heap data structure is adopted, which is similar to the implementation by Lau \textit{et al.} \cite{lau2010improved}. This cache-oblivious data structure minimizes the I/O operations in theory. Following the work of \cite{iacono2019parallel}, we have also implemented a GPU version\footnote{https://github.com/JINXER000/parHeap} of the bucket heap, aiming to speed up the computation. However, it turns out that this data structure is not suitable for GPU since its performance is even worse than that of the CPU bucket heap. In contrast, our proposed method is cache-aware and we have the freedom to decide the parameters catering to different computing platforms. Other technqiues included for comparison with our works are the parallel Bellman-Ford method \cite{busato2015efficient}, the BFS (breadth-first search) method accelerated by prefix-sum\footnote{https://github.com/rafalk342/bfs-cuda}, and the algorithm with GPU pripority queue\footnote{https://github.com/crosetto/cupq}.

\begin{table}[htbp]
		\centering
		\footnotesize
		\caption{Global EDT integration time with different propagation algorithms}
		\label{tab:diff-prop}
		\tabcolsep 3pt %
		\renewcommand\arraystretch{1.2}
		\begin{threeparttable}
\begin{tabular}{c c}
\hline \hline
\textbf{Method}          & \textbf{Average Updating Time (ms)}  \\\hline 
CPU bucket heap & 790 \\
GPU bucket heap & 2035 \\
GPU priority queue & 3020 \\
GPU Bellman-Ford & 250 \\
GPU Prefix Sum BFS & 22 \\
Ours & \textbf{10} \\ \hline \hline
\end{tabular}
\end{threeparttable}
\end{table}

The following experiments are conducted to show that our method outperforms the state-of-the-art EDT mapper, FIESTA \cite{han2019fiesta}. All  comparisons are done with the UGV-Corridor dataset, and the local volume is set to 10~m~$\times$~10~m~$\times$~1.2~m. For results on the Cow-and-Lady dataset, we refer interested readers to the supplementary video.

\begin{figure}[htb]
\vspace{-0.6cm}
		\subfloat[OGM construction time on PC]
{\includegraphics[width=0.22\textwidth]{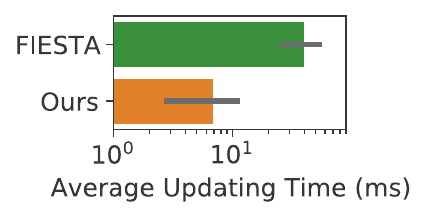} \label{fig:occubarplot1}
}\hfill
		\subfloat[OGM construction time on Xavier NX]
{\includegraphics[width=0.22\textwidth]{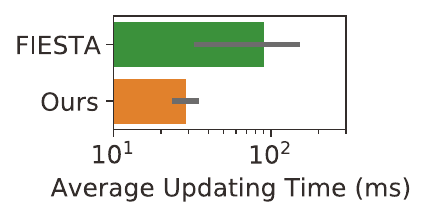} \label{fig:occubarplot3}
}\hfill

		\subfloat[EDT time on PC]
{\includegraphics[width=0.22\textwidth]{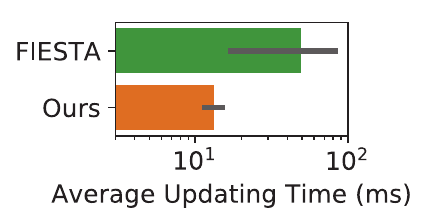} \label{fig:sdfbarplot1}
}\hfill
		\subfloat[EDT time on Xavier NX]
{\includegraphics[width=0.22\textwidth]{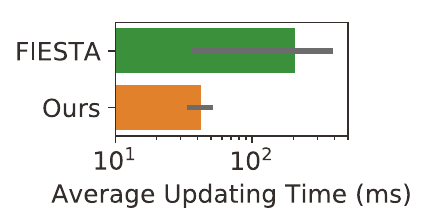} \label{fig:sdfbarplot3}
}\hfill
	\caption{The mapping results of different mappers on different platforms. The bold line in each center of the bar is the standard deviation of time throughout the run. }
	\label{fig:barplot}
\end{figure}

First, we evaluate the performance of mappers on different computing platforms, i.e., a personal computer (with Intel i5-8300H CPU and a GTX1060 GPU) and an Nvidia Xavier NX. {Since  high-frequency local planning is usually limited within a local volume,  the CPU-GPU stream is turned off in our method, while the accurate EDT inside the local volume is copied to CPU and broadcast  by ROS (although the global EDT in our method can be directly accessed by a GPU-based motion planner \cite{lu2021flight}).  In FIESTA, we simply disable the \textit{global\_update} option to update its EDT only within the local volume, which makes the condition work in its favor.}   The frequency of updating the EDT is 2~Hz, and the resolution is 0.05~m. 
{The elapsed time of constructing OGM and EDT are measured separately, and the results are plotted in bar charts as shown in Fig.~\ref{fig:barplot}} .
Obviously,  our method outperforms FIESTA in speed on both platforms.
Furthermore, as discussed in Section~\ref{sec changerate}, the average EDT time of our method is expected to be less volatile.

\begin{figure}[htbp]  %
    \centering    %
        \subfloat[RMS Error w.r.t. Voxel Size] %
    { \label{fig:rms-s}
        \begin{minipage}[t]{0.25\textwidth}
            \centering      %
            \includegraphics[width=1\textwidth]{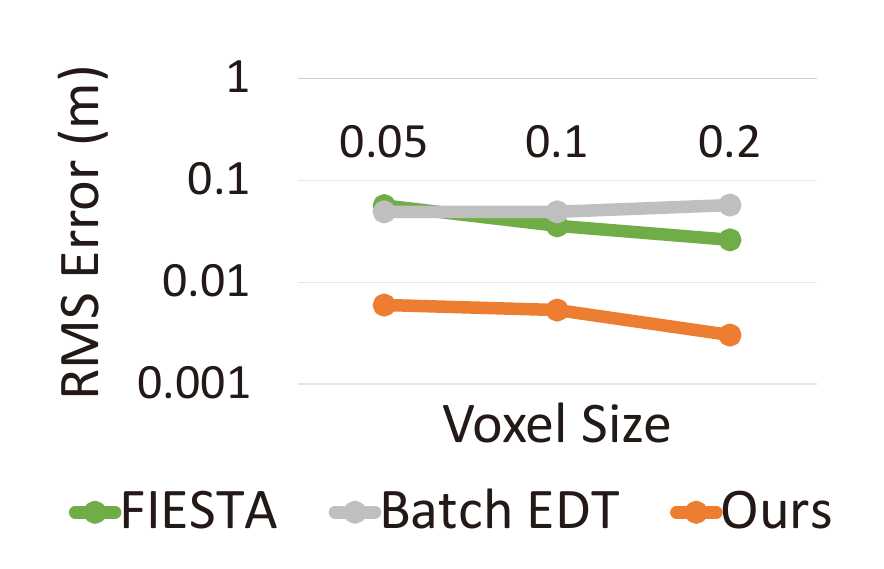}   %
          
        \end{minipage}
    }%
    \subfloat[RMS Error  w.r.t. Connectivity] %
    { \label{fig:rms-c}
        \begin{minipage}[t]{0.24\textwidth}
            \centering          %
            \includegraphics[width=1\textwidth]{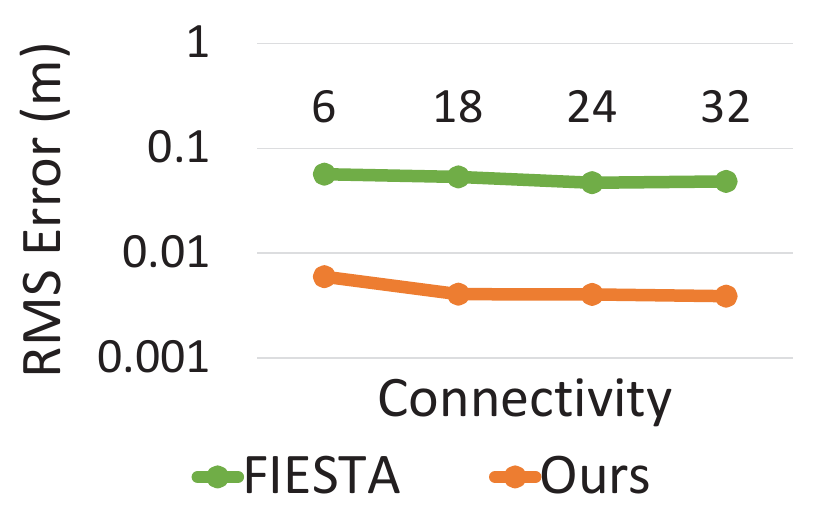}   %
           
        \end{minipage}%
    }%

    \caption{The RMS error in different parameter settings. } %
    \label{fig:accuracy}
\end{figure}

The second set of experiments is performed to investigate the extent to which hyperparameters, such as voxel size and connectivity, affect the accuracy of the EDT within the local volume. The results are shown in Fig.~\ref{fig:accuracy}.
The mappers in this experiment take measurements from the 
UGV-Corridor dataset, construct their own OGM and EDT, and then calculate the RMS error of the EDT  against  ground truth.
In the ground truth calculation, a K-D tree of the current OGM constructed by each mapper is first created. The nearest neighbor of each voxel is then found and the least distance to the obstacles is computed.  Note that we only focus on the error induced by the EDT computation and do not include the error caused by the discretization of OGM. 
In Fig.~\subref*{fig:rms-s}, all samples are tested with 6-connected in propagation. 
{Since FIESTA is more accurate than  Voxblox \cite{oleynikova2017voxblox} according to Han \textit{et al.}  \cite{han2019fiesta}, we only compare our method with  FIESTA. In both FIESTA and our method, global update is enabled while only the error in the local volume is taken into consideration. Also, the RMS error curve of the batch EDT version of our mapper is plotted to show the necessity of global EDT integration. 
We can see that the resolution and the RMS error are negatively correlated. 
The mapping accuracy with respect to connectivity is shown in  Fig.~\subref*{fig:rms-c}, where the resolution is fixed as  0.05~m.} As the connectivity increases, the RMS error drops. From similar results in FIESTA, a likely explanation is that the error  is mainly incurred by wavefront propagation. When the connectivity remains constant, as the number of voxels increases, error cases in the propagation has a larger chance to occur. On the other hand, as more neighbors are considered,  the  propagation error could be reduced. We refer interested readers to \cite{cuisenaire1999fast} for a more descriptive explanation.
Compared to FIESTA, our mapper is more accurate in any situation because the exactness of our batch EDT method helps to reduce the error. 

\vspace{-0.5cm}

\begin{figure}[htbp]  %
    \centering    %
        \subfloat[Global EDT time on Xavier NX] %
    { \label{fig:glb-spd}
        \begin{minipage}[t]{0.24\textwidth}
            \centering      %
            \includegraphics[width=1\textwidth]{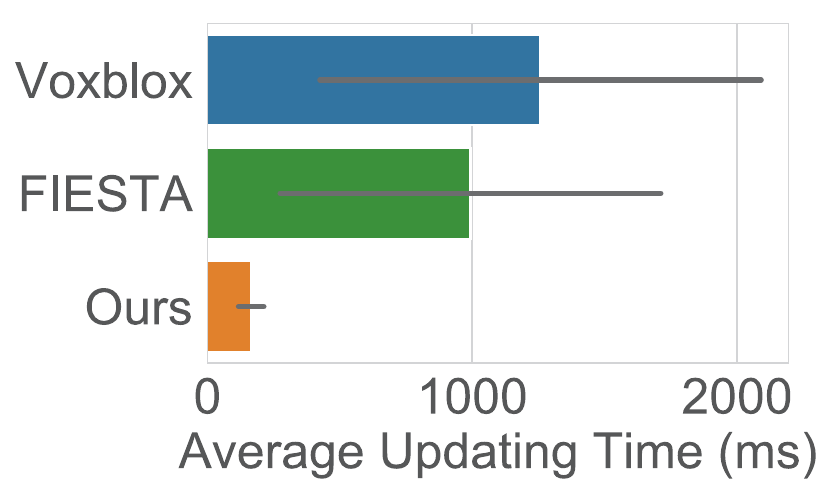}   %
        \end{minipage}
    }%
    \subfloat[Global RMS Error  w.r.t. Voxel Size] %
    { \label{fig:glb-acc}
        \begin{minipage}[t]{0.24\textwidth}
            \centering          %
            \includegraphics[width=1\textwidth]{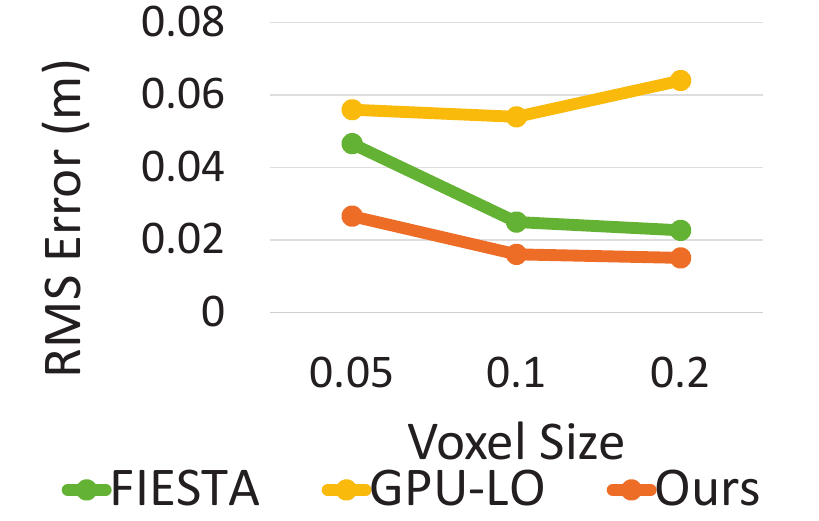}   %
        \end{minipage}%
    }%

    \caption{The average speed and RMS error measured in the entire map. } %
    \label{fig:glb-spd-acc}
\end{figure}

{In addition to the experiments in the local volume, we compared the mapping performance  with Voxblox \cite{oleynikova2017voxblox}, FIESTA \cite{han2019fiesta}, and GPU-LO \cite{chen2021gpu} in terms of global EDT. From Fig~\subref*{fig:glb-spd}, although maintaining the global EDT on both GPU and CPU causes extra overhead in our method, it is negligible compared to the drastically increasing load in FIESTA. Similarly, as shown in Fig~\subref*{fig:glb-acc}, when measuring RMS error over the whole map,  the newly proposed technique has outperformed FIESTA and GPU-LO.  The results assure us that our system suffices to provide  global EDT  for online global motion planners.   }

\section{Implementation on Actual Robotic Platforms}
\label{newsection}
\begin{figure}[htb]
		\subfloat[]
{\includegraphics[width=0.23\textwidth]{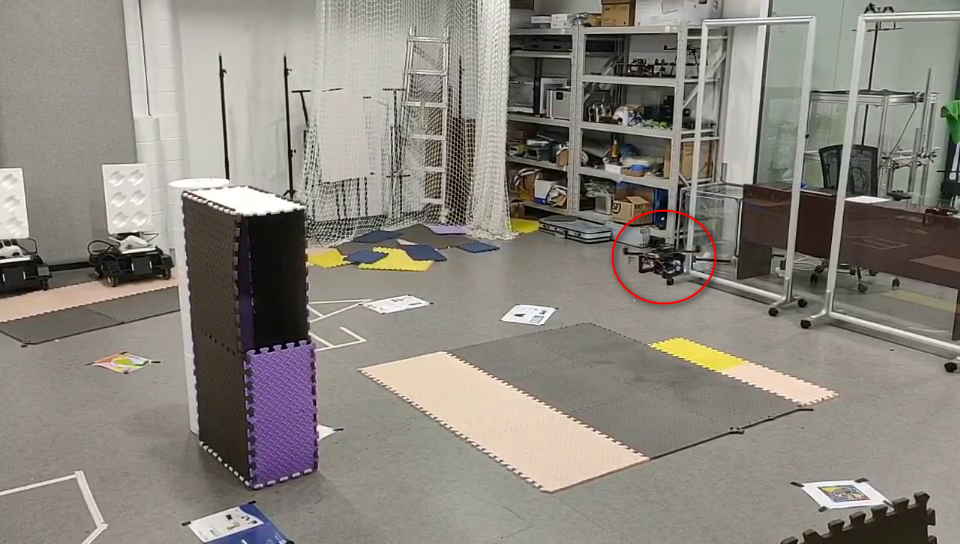} \label{fig:uav-test3}
}\hfill
		\subfloat[]
{\includegraphics[width=0.23\textwidth]{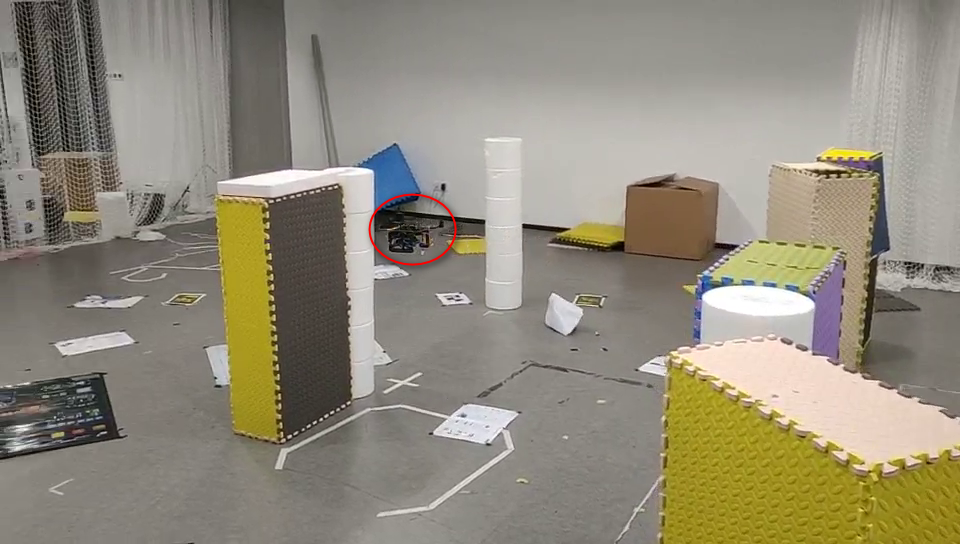} \label{fig:uav-test4}
}\hfill
		\subfloat[]
{\includegraphics[width=0.23\textwidth]{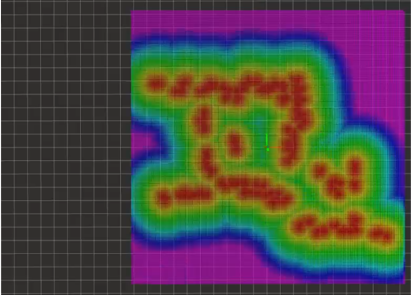} \label{fig:uav-test1}
}\hfill
		\subfloat[]
{\includegraphics[width=0.23\textwidth]{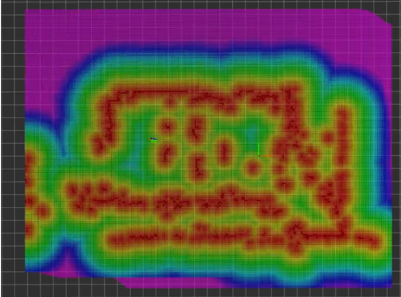} \label{fig:uav-test2}
}\hfill
	\caption{ The snapshots and the constructed maps when an MAV  flying through obstacles. The movement is limited to a 2D plane since only a 2D LiDAR is used. }
	\label{fig:edt-uav}
\end{figure}

To verify that the entire mapping system can run on lightweight onboard computers, we conduct an experiment with an MAV navigating in an unexplored and cluttered indoor environment. The motion planner is taken directly from \cite{lai2019model}, which requires  distance information from EDT. The range measurement comes from a 2D LiDAR mounted on the top of the MAV. An Intel RealSense T-265 tracking camera is used to estimate the current pose of the MAV. All computations are done on an onboard Nvidia Xavier NX. In Fig.~\ref{fig:edt-uav}, the MAV flies to an unmapped region while avoiding obstacles. {The mapper updates EDT at 10~Hz within 20~m~$\times$~20~m with 0.1~m voxel width, allowing the motion planner to obtain the required distance information in real-time.}  
\vspace{-0.3cm}

\begin{figure*}[htb]
	\centering
	\includegraphics[width=0.7\textwidth]{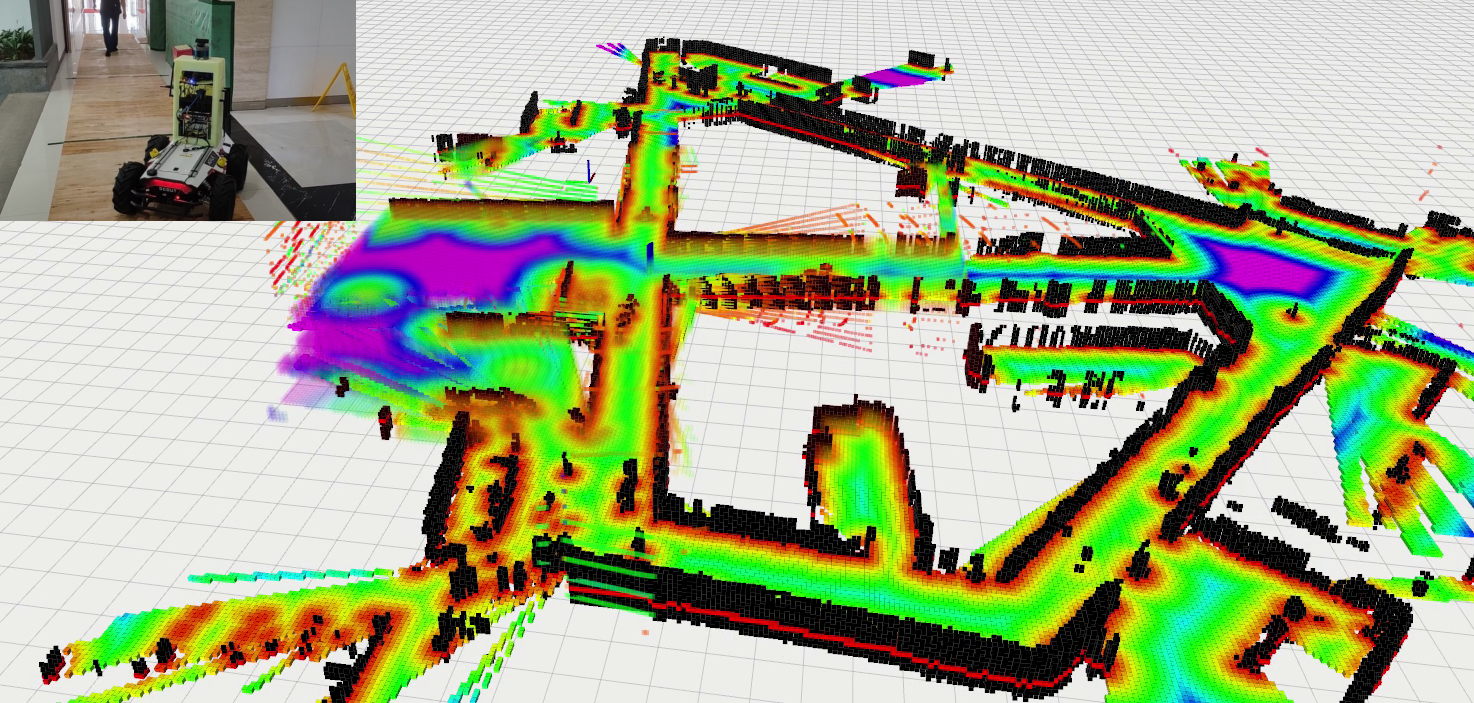}
	\caption{OGM and EDT constructed by a UGV in  narrow corridors.}
	\label{fig:1st-pic}
\end{figure*}

Lastly, Fig.~\ref{fig:1st-pic} depicts a global EDT map constructed using a UGV equipped with a 16-line LiDAR. The experiment was carried out in an indoor environment with narrow corridors. We set the resolution as 0.1~m while the local volume is 20~m~$\times$~20~m$\times$~1.2~m.  Using a computer with Intel i5 9300H CPU and Nvidia 1660Ti GPU, the vehicle navigates fully autonomously through walking pedestrians. The results show that our proposed technique is efficient and can be readily adopted for real applications. 

\section{Conclusion} \label{Last_Sec}

In this letter, we  have proposed a novel robotic volumetric mapping framework with parallel computing. The proposed system is capable of generating real-time OGMs and EDTs that can be used for rapid motion planning. The effectiveness and the efficiency of our proposed technique has been proved by various experiments in simulations, datasets, and the real world problems. With the deployment on the onboard computing devices of both MAV and UGV, we have demonstrated that our system can efficiently build incremental EDT for mobile robots to navigate in unexplored environments.

\bibliographystyle{IEEEtran}   %
\bibliography{IEEEabrv,bib/IEEEexample}   %

\begin{thebibliography}{10}
\providecommand{\url}[1]{#1}
\csname url@samestyle\endcsname
\providecommand{\newblock}{\relax}
\providecommand{\bibinfo}[2]{#2}
\providecommand{\BIBentrySTDinterwordspacing}{\spaceskip=0pt\relax}
\providecommand{\BIBentryALTinterwordstretchfactor}{4}
\providecommand{\BIBentryALTinterwordspacing}{\spaceskip=\fontdimen2\font plus
\BIBentryALTinterwordstretchfactor\fontdimen3\font minus
  \fontdimen4\font\relax}
\providecommand{\BIBforeignlanguage}[2]{{%
\expandafter\ifx\csname l@#1\endcsname\relax
\typeout{** WARNING: IEEEtran.bst: No hyphenation pattern has been}%
\typeout{** loaded for the language `#1'. Using the pattern for}%
\typeout{** the default language instead.}%
\else
\language=\csname l@#1\endcsname
\fi
#2}}
\providecommand{\BIBdecl}{\relax}
\BIBdecl

\bibitem{lan2021survey}
M.~Lan, S.~Lai, T.~H. Lee, and B.~M. Chen, ``A survey of motion and task
  planning techniques for unmanned multicopter systems,'' \emph{Unmanned
  Systems}, vol.~9, no.~02, pp. 165--198, 2021.

\bibitem{mondal2021offline}
M.~Mondal and S.~Poslavskiy, ``Offline navigation (homing) of aerial vehicles
  (quadcopters) in {GPS} denied environments,'' \emph{Unmanned Systems},
  vol.~9, no.~02, pp. 119--127, 2021.

\bibitem{yuan2021survey}
S.~Yuan, H.~Wang, and L.~Xie, ``Survey on localization systems and algorithms
  for unmanned systems,'' \emph{Unmanned Systems}, vol.~9, no.~02, pp.
  129--163, 2021.

\bibitem{cui2016search}
J.~Q. Cui, S.~K. Phang, K.~Z. Ang, F.~Wang, X.~Dong, Y.~Ke, S.~Lai, K.~Li,
  X.~Li, J.~Lin \emph{et~al.}, ``Search and rescue using multiple drones in
  post-disaster situation,'' \emph{Unmanned Systems}, vol.~4, no.~01, pp.
  83--96, 2016.

\bibitem{ratliff2009chomp}
N.~Ratliff, M.~Zucker, J.~A. Bagnell, and S.~Srinivasa, ``Chomp: Gradient
  optimization techniques for efficient motion planning,'' in \emph{2009 IEEE
  International Conference on Robotics and Automation}.\hskip 1em plus 0.5em
  minus 0.4em\relax IEEE, 2009, pp. 489--494.

\bibitem{zhou2019robust}
B.~Zhou, F.~Gao, L.~Wang, C.~Liu, and S.~Shen, ``Robust and efficient quadrotor
  trajectory generation for fast autonomous flight,'' \emph{IEEE Robotics and
  Automation Letters}, vol.~4, no.~4, pp. 3529--3536, 2019.

\bibitem{arora2021environment}
P.~Arora and C.~Papachristos, ``Environment reconfiguration planning for
  autonomous robotic manipulation to overcome mobility constraints,'' in
  \emph{2021 IEEE International Conference on Robotics and Automation
  (ICRA)}.\hskip 1em plus 0.5em minus 0.4em\relax IEEE, 2021, pp. 2352--2358.

\bibitem{jarvis1993distance}
R.~Jarvis, ``Distance transform based path planning for robot navigation,'' in
  \emph{Recent trends in mobile robots}.\hskip 1em plus 0.5em minus 0.4em\relax
  World Scientific, 1993, pp. 3--31.

\bibitem{lau2010improved}
B.~Lau, C.~Sprunk, and W.~Burgard, ``Improved updating of {Euclidean} distance
  maps and {Voronoi} diagrams,'' in \emph{2010 IEEE/RSJ International
  Conference on Intelligent Robots and Systems}.\hskip 1em plus 0.5em minus
  0.4em\relax IEEE, 2010, pp. 281--286.

\bibitem{oleynikova2017voxblox}
H.~Oleynikova, Z.~Taylor, M.~Fehr, R.~Siegwart, and J.~Nieto, ``Voxblox:
  Incremental {3D} {Euclidean} signed distance fields for on-board {MAV}
  planning,'' in \emph{2017 Ieee/rsj International Conference on Intelligent
  Robots and Systems (iros)}.\hskip 1em plus 0.5em minus 0.4em\relax IEEE,
  2017, pp. 1366--1373.

\bibitem{han2019fiesta}
L.~Han, F.~Gao, B.~Zhou, and S.~Shen, ``Fiesta: Fast incremental euclidean
  distance fields for online motion planning of aerial robots,'' in \emph{2019
  IEEE/RSJ International Conference on Intelligent Robots and Systems
  (IROS)}.\hskip 1em plus 0.5em minus 0.4em\relax IEEE, 2019, pp. 4423--4430.

\bibitem{saito1994new}
T.~Saito and J.-I. Toriwaki, ``New algorithms for {Euclidean} distance
  transformation of an n-dimensional digitized picture with applications,''
  \emph{Pattern recognition}, vol.~27, no.~11, pp. 1551--1565, 1994.

\bibitem{meijster2002general}
A.~Meijster, J.~B. Roerdink, and W.~H. Hesselink, ``A general algorithm for
  computing distance transforms in linear time,'' in \emph{Mathematical
  Morphology and its applications to image and signal processing}.\hskip 1em
  plus 0.5em minus 0.4em\relax Springer, 2002, pp. 331--340.

\bibitem{Cao2010Parallel}
T.~T. Cao, K.~Tang, A.~Mohamed, and T.~S. Tan, ``Parallel banding algorithm to
  compute exact distance transform with the {GPU},'' in \emph{Acm Siggraph
  Symposium on Interactive {3D} Graphics and Games}, 2010.

\bibitem{vespa2018efficient}
E.~Vespa, N.~Nikolov, M.~Grimm, L.~Nardi, P.~H. Kelly, and S.~Leutenegger,
  ``Efficient octree-based volumetric slam supporting signed-distance and
  occupancy mapping,'' \emph{IEEE Robotics and Automation Letters}, vol.~3,
  no.~2, pp. 1144--1151, 2018.

\bibitem{funk2021multi}
N.~Funk, J.~Tarrio, S.~Papatheodorou, M.~Popovi{\'c}, P.~F. Alcantarilla, and
  S.~Leutenegger, ``Multi-resolution {3D} mapping with explicit free space
  representation for fast and accurate mobile robot motion planning,''
  \emph{IEEE Robotics and Automation Letters}, vol.~6, no.~2, pp. 3553--3560,
  2021.

\bibitem{wu2021faithful}
L.~Wu, K.~M.~B. Lee, L.~Liu, and T.~Vidal-Calleja, ``Faithful {Euclidean}
  distance field from log-{Gaussian} process implicit surfaces,'' \emph{IEEE
  Robotics and Automation Letters}, vol.~6, no.~2, pp. 2461--2468, 2021.

\bibitem{niessner2013real}
M.~Nie{\ss}ner, M.~Zollh{\"o}fer, S.~Izadi, and M.~Stamminger, ``Real-time {3D}
  reconstruction at scale using voxel hashing,'' \emph{ACM Transactions on
  Graphics (ToG)}, vol.~32, no.~6, pp. 1--11, 2013.

\bibitem{chen2021gpu}
Y.~Chen, S.~Lai, B.~Wang, F.~Lin, and B.~M. Chen, ``A {GPU} mapping system for
  real-time robot motion planning,'' in \emph{2021 IEEE International
  Conference on Real-time Computing and Robotics (RCAR)}.\hskip 1em plus 0.5em
  minus 0.4em\relax IEEE, 2021, pp. 762--768.

\bibitem{amanatides1987fast}
J.~Amanatides, A.~Woo \emph{et~al.}, ``A fast voxel traversal algorithm for ray
  tracing,'' in \emph{Eurographics}, vol.~87, no.~3, 1987, pp. 3--10.

\bibitem{curless1996volumetric}
B.~Curless and M.~Levoy, ``A volumetric method for building complex models from
  range images,'' 1996.

\bibitem{hynninen2017cutt}
A.-P. Hynninen and D.~I. Lyakh, ``{cuTT}: A high-performance tensor transpose
  library for {CUDA} compatible {GPU}s,'' \emph{arXiv preprint
  arXiv:1705.01598}, 2017.

\bibitem{harish2007accelerating}
P.~Harish and P.~J. Narayanan, ``Accelerating large graph algorithms on the
  {GPU} using {CUDA},'' in \emph{International conference on high-performance
  computing}.\hskip 1em plus 0.5em minus 0.4em\relax Springer, 2007, pp.
  197--208.

\bibitem{deng2009taming}
Y.~Deng, B.~D. Wang, and S.~Mu, ``Taming irregular eda applications on
  {GPU}s,'' in \emph{Proceedings of the 2009 International Conference on
  Computer-Aided Design}, 2009, pp. 539--546.

\bibitem{luo2010effective}
L.~Luo, M.~Wong, and W.-m. Hwu, ``An effective {GPU} implementation of
  breadth-first search,'' in \emph{Design Automation Conference}.\hskip 1em
  plus 0.5em minus 0.4em\relax IEEE, 2010, pp. 52--55.

\bibitem{iacono2019parallel}
J.~Iacono, B.~Karsin, and N.~Sitchinava, ``A parallel priority queue with fast
  updates for {GPU} architectures,'' \emph{arXiv preprint arXiv:1908.09378},
  2019.

\bibitem{busato2015efficient}
F.~Busato and N.~Bombieri, ``An efficient implementation of the {Bellman-Ford}
  algorithm for {Kepler} {GPU} architectures,'' \emph{IEEE Transactions on
  Parallel and Distributed Systems}, vol.~27, no.~8, pp. 2222--2233, 2015.

\bibitem{lu2021flight}
H.~Lu, Q.~Zong, S.~Lai, B.~Tian, and L.~Xie, ``Flight with limited field of
  view: A parallel and gradient-free strategy for micro aerial vehicle,''
  \emph{IEEE Transactions on Industrial Electronics}, 2021.

\bibitem{cuisenaire1999fast}
O.~Cuisenaire and B.~Macq, ``Fast {Euclidean} distance transformation by
  propagation using multiple neighborhoods,'' \emph{Computer vision and Image
  understanding}, vol.~76, no.~2, pp. 163--172, 1999.

\bibitem{lai2019model}
S.~Lai, M.~Lan, and B.~M. Chen, ``Model predictive local motion planning with
  boundary state constrained primitives,'' \emph{IEEE Robotics and Automation
  Letters}, vol.~4, no.~4, pp. 3577--3584, 2019.

\end{thebibliography}

\end{document}